# VISION-BASED NAVIGATION OF AUTONOMOUS VEHICLE IN ROADWAY ENVIRONMENTS WITH UNEXPECTED HAZARDS

**Mhafuzul Islam***
**Ph.D. Student**
Glenn Department of Civil Engineering, Clemson University
351 Fluor Daniel Engineering Innovation Building, Clemson, SC 29634
Tel: (864) 986-5446, Fax: (864) 656-2670
Email: mdmhafi@clemson.edu

**Mashrur Chowdhury, Ph.D., P.E., F. ASCE**
**Eugene Douglas Mays Professor of Transportation**
Glenn Department of Civil Engineering, Clemson University
216 Lowry Hall, Clemson, South Carolina 29634
Tel: (864) 656-3313, Fax: (864) 656-2670
E-mail: mac@clemson.edu

**Hongda Li**
**Ph.D. Student**
Division of Computer Science
School of Computing, Clemson University
220 McAdams Hall, Clemson, South Carolina 29634
Tel: (864) 986-2018
E-mail: hongdal@clemson.edu

**Hongxin Hu, Ph.D.**
**Assistant Professor**
Division of Computer Science
School of Computing, Clemson University
217 McAdams Hall, Clemson, South Carolina 29634
Tel: (864) 656-2847
E-mail: hongxih@clemson.edu

**Submitted to Transportation Research Record - 2019**

**ABSTRACT**

Vision-based navigation of autonomous vehicles primarily depends on the Deep Neural Network (DNN) based systems in which the controller obtains input from sensors/detectors, such as cameras and produces a vehicle control output, such as a steering wheel angle to navigate the vehicle safely in a roadway traffic environment. Typically, these DNN-based systems of the autonomous vehicle are trained through supervised learning; however, recent studies show that a trained DNN-based system can be compromised by perturbation or adversarial inputs. Similarly, this perturbation can be introduced into the DNN-based systems of autonomous vehicle by unexpected roadway hazards, such as debris and roadblocks. In this study, we first introduce a roadway hazardous environment (both intentional and unintentional roadway hazards) that can compromise the DNN-based navigational system of an autonomous vehicle, and produces an incorrect steering wheel angle, which can cause crashes resulting in fatality and injury. Then, we develop a DNN-based autonomous vehicle driving system using object detection and semantic segmentation to mitigate the adverse effect of this type of hazardous environment, which helps the autonomous vehicle to navigate safely around such hazards. We find that our developed DNN-based autonomous vehicle driving system including hazardous object detection and semantic segmentation improves the navigational ability of an autonomous vehicle to avoid a potential hazard by 21% compared to the traditional DNN-based autonomous vehicle driving system.

## INTRODUCTION

According to the 2016 American automobile association report, 50,658 crashes occurred in the U.S. from the year 2011 to 2014 due to roadway hazards resulting in 9,805 injuries and 125 deaths (*1*). The roadway hazards, such as debris, are considered to be non-fixed and unexpected objects on the travel or driving lane of the roadway and include objects that have fallen from vehicles or have come from construction sites or littering. Given that the autonomous vehicle is considered as the future of surface transportation, its ability to detect debris or hazards and then navigate safely around them is crucial for avoiding potential crashes. Recently, such navigational task has been accomplished using Deep Neural Network (DNN). Typically, an autonomous vehicle perceives its surrounding roadway environment using sensors, and the software running in the vehicle determines the action to be taken based on the input from the sensors. Several types of sensors, such as vision-based sensor (e.g., Camera), LIDAR, and Radar are currently available for the perception task. Due to the cost-effectiveness of the vision-based sensor compared to the other types of sensors (e.g., LIDAR and Radar), vision-based navigation becomes an attractive solution for autonomous vehicles (*2*)(*3*)(*4*).

The recent development of DNNs, in particular, Convolutional Neural Network (CNN)(*5*), has improved vision-based navigation for autonomous vehicles significantly. After being trained and tested using a dataset collected by sensors, these CNN models are then deployed in autonomous vehicles to navigate the vehicle safely. For example, during training, the CNN-based end-to-end driving model maps a relationship between the driving behavior of humans using roadway images collected from cameras and the steering wheel angle (*6*)(*7*). Thus, the performance of autonomous vehicles primarily depends on the training dataset, meaning if a hazard that the CNN model is not trained on appears on the roadway, the autonomous vehicle driving model may produce an incorrect steering wheel angle and may cause a crash. A recent study shows that the autonomous vehicle navigation system may fail to navigate safely due to several reasons, such as Radar sensor failure, camera sensor failure, and software failure (*8*). This study addresses the situation where a well-trained driving model may fail due to unexpected hazards that may lead to unsafe navigation, and then explores the use of object detection and semantic segmentation (*9*) for mitigating the navigational problem in this hazardous condition.

The remainder of the paper is organized as follows. The related work section explores the existing studies on autonomous vehicle navigation, state-of-art DNN-based autonomous vehicle driving models, and the limitations of the traditional DNN-based model. Then we introduce the method developed in this study for navigating an autonomous vehicle on a roadway with unexpected hazards. Furthermore, we validate our proposed method using three case studies: (i) a model trained using a dataset that includes hazards but without considering them as separate input features; (ii) a model trained on a dataset that considers hazards as separate input features and uses a distance measurement sensor and image segmentation; (iii) a model trained on a dataset that considers hazards as separate input features and only uses image segmentation. In the second and third case studies, we introduce a DNN-based autonomous vehicle driving system to enhance the ability of an autonomous vehicle to navigate safely in a hazardous environment. Then we present the experimental setup employed in this study. After that, we evaluate all the case scenarios and report the results obtained through our experiments, and finally, we discuss the conclusions and suggest the areas for future work.

## RELATED WORK

This section reviews the previous research on hazard detection, DNN-based driving systems used in an autonomous vehicle, and the techniques for and the importance of object detection and image segmentation in addition to the limitations of using DNN in autonomous vehicles.

### DNN-based Autonomous Vehicle Driving Model

DNN-based autonomous vehicle driving systems are rapidly evolving (*7*)(*10*). Not only software companies such as Waymo (Google) Uber, and Lyft are using the DNN-based systems for autonomous vehicles, but many car companies such as Tesla, Volvo, BMW, and Ford are currently working on DNN-based autonomous vehicle driving systems (*11*). In such systems, sensors like cameras, LIDAR, and Radar provide input to DNN models, such as Convolutional Neural Network (CNN)(*5*) or Recurrent Neural Network (RNN) (*12*), which then produce outputs such as steering wheel angle and velocity. For example, the autonomous vehicle architecture developed by NVIDIA, named DAVE-2, uses a CNN model which takes input from a camera and outputs a steering wheel commands for navigation (*7*), while Udacity autonomous vehicle driving architectures include both CNN-based (e.g., Autumn) and RNN-based (e.g., Chauffeur using CNN and RNN) (*13*). This study used a CNN-based driving model similar to DAVE-2 as it is the fundamental base of DNN-based autonomous vehicle systems.

DNN-based autonomous vehicle driving systems, which are intrinsically software systems, can be error-prone and cause severe consequences if they do not function as intended. Several studies have shown the vulnerabilities of the existing DNN models (*14*)(*15*)(*16*)(*17*). For example, DNN-based image classification can be exploited by adding a small perturbation to an input image such that the DNN model misclassifies it as another category, a vulnerability recently confirmed by (*18*), which found that attackers can physically modify objects using a low-cost technique to cause classification errors in DNN-based vision systems. These perturbations can be introduced under widely varying distances, angles, and resolutions. For example, in (*18*) perturbations caused a DNN model to interpret a subtly modified physical stop sign as a speed limit of 45 mph sign. Similarly, the debris or roadblocks on the road can also compromise the autonomous vehicle driving system by producing incorrect steering wheel angles, potentially causing a fatal collision. These limitations prompted this study to evaluate the impact of unexpected hazardous environments on a DNN-based autonomous vehicle driving system.

### Autonomous Vehicle Dataset

Data are an important part of deep learning-based systems, and this study requires a dataset that supports (i) end-to-end driving systems (input: image; output: steering wheel angle), (ii) image segmentation, and (iii) hazard detection. To find an appropriate one, we explore various existing datasets used by the autonomous vehicle community. The closest dataset provided by Udacity, which supports end-to-end data and image segmentation, but it does not provide the ground truth for hazards in the drivable lane (*13*). KTTI (*19*) and Cityscape (*20*) datasets also do not support hazard detection as ground truth data. The dataset matching our requirements the closest is the Lost and Found dataset (*21*), which contains the image as the input, and the yaw rate (angular velocity), but not the steering wheel angle required by this study, as an output. Since existing datasets do not fully meet our needs, after careful consideration, we have created our own dataset using simulation as described in the experimental setup section.

**DNN-based Object Detection and Segmentation**

Object detection and classification are core components of autonomous driving. By detecting and classifying the objects, the autonomous vehicle controller determines safe navigation for both path planning and route planning. If an autonomous vehicle is not able to detect unexpected hazards on the road, it will not be able to navigate safely, perhaps resulting in a crash. However, detecting these objects or hazards is a challenging task. While various sensors, such as Radar and LIDAR, can be used for accurate distance and velocity measurement, these sensors are relatively costly than camera sensor (*21*). Considering these limitations, vision-based sensors, such as camera, are being used on autonomous vehicles for the navigational task. With the recent development of DNNs, DNN-based object detection and semantic segmentation can be applied to detect these roadway hazards, making navigation of autonomous vehicles safer.

Semantic segmentation is a technology that has been widely used in the computer vision area to divide an unknown image into different parts (*22*), can be applied to an image containing unknown objects. This technology is effective in providing the scenario depicted by an image, allowing the DNN to capture additional information about the dataset during training. There are three major types of semantic segmentation technologies: Region-based semantic segmentation (*23*)(*24*), Fully Convolutional Network (FCN)-based semantic segmentation (*25*)(*26*)(*27*) and Weakly-Supervised semantic segmentation (*28*)(*29*)(*30*). The region-based semantic segmentation provides segmentation based on the results of object detection, meaning it can be developed on any CNN model. The FCN-based semantic segmentation segments each pixel of the image, meaning it does not require extracting regions of the image and, thus, can be applied to arbitrary sizes of images. The weakly supervised semantic segmentation technology, which was developed to reduce the labeling cost of a large dataset (*30*), achieves semantic segmentation by exploiting annotated bounding boxes or image-level labels. While recent studies show that segmentation-based navigation can improve navigational performance (*31*)(*32*)(*33*), none considers the navigation of autonomous vehicles in hazardous environments. Thus, by leveraging these DNN-based models, we can detect hazards and then extract their semantic information from images obtained from the camera sensor of an autonomous vehicle. The approach adopted in this study uses an FCN-based model as one such network is relatively small, yet the network yields fast results (*25*). To the best of our knowledge, this is the first work that develops a DNN-based autonomous vehicle driving system focusing on unexpected roadway hazardous environments.

## METHOD

In this section, we describe our approach for developing a safer autonomous vehicle driving system in a hazardous environment. This study uses DNN-based object detection and segmentation to create a corrected image, which is subsequently used by the autonomous vehicle driving system to predict the steering wheel angle. As presented in Figure 1, we develop a DNN-based autonomous vehicle driving system, which comprises of three DNN models. The first one is the DNN-based hazard detection and segmentation model, which detects the hazard and creates a segmented image. The second model is the hazard analysis and avoidance model, which fuses the segmented image with the original input image from the dashboard camera to make the autonomous vehicle driving model aware of the unexpected roadway hazards. This model then analyzes the hazard and determines if the hazard should be ignored or considered as a threat for a potential crash using a threat factor ($T_f$). The third model is the DNN-based autonomous vehicle driving model, which takes the fused image with hazard information and produces the steering wheel angle required to navigate the vehicle safely in an unexpected hazardous environment. We provide the detail description of these three models in the following subsections.

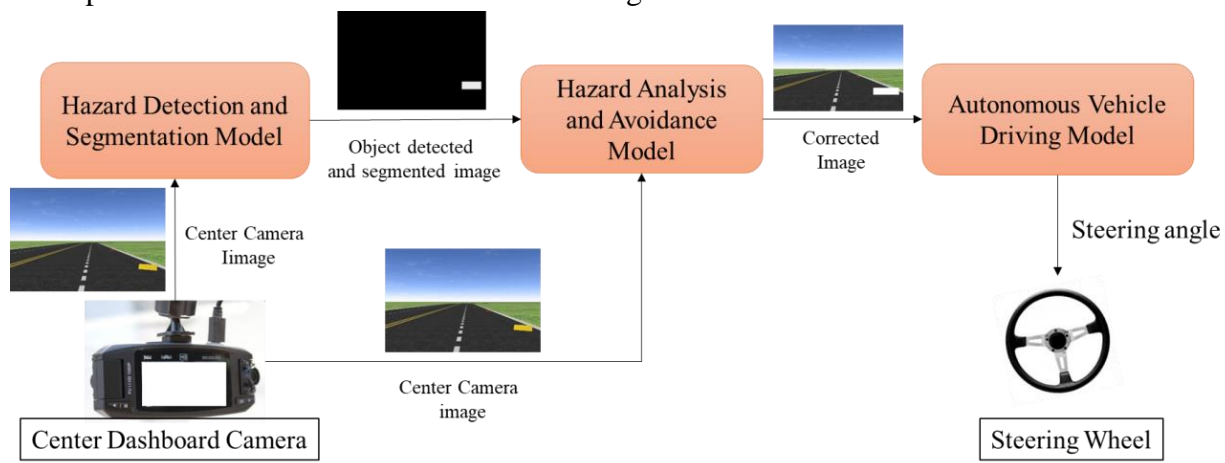

**FIGURE 1 DNN-based autonomous vehicle driving system in an unexpected hazardous environment.**

### DNN-based hazard detection and segmentation model

For hazard detection and image segmentation, this study uses an FCN, which is a DNN-based image object detection and segmentation model (*25*). Figure 2 shows the structure of the FCN network used in our study. It takes an input of image size 400x600x3 and outputs a segmented image of the same size. We use a pre-trained network with a weight of VGGNet (*34*), which is a deep convolutional network for large-scale image recognition, and then we re-trained the model with our training dataset to classify the hazard and perform image segmentation.

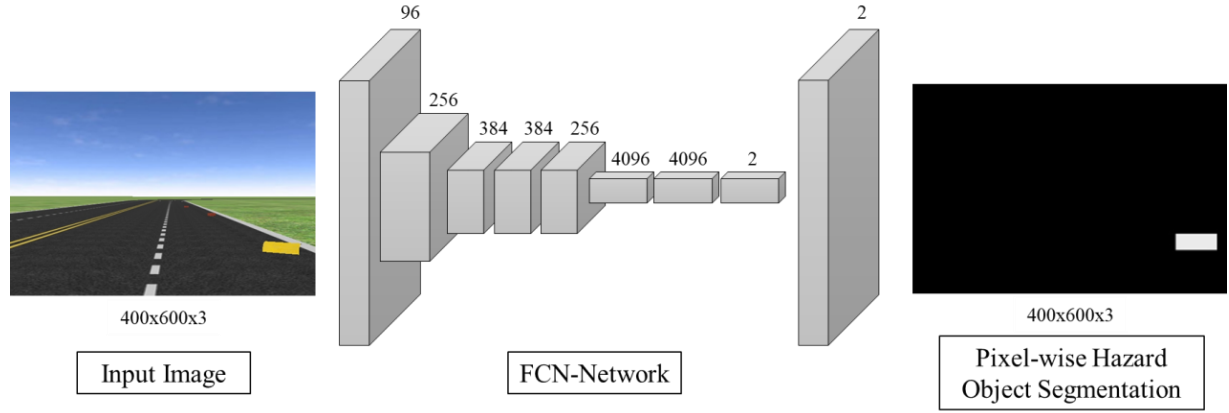

**FIGURE 2 FCN-based object detection and image segmentation model used in this study.**

## Hazard analysis and avoidance model

As shown in Figure 1, the image captured from the center dashboard camera first goes to the hazard detection and segmentation model, which provides an output of the detected object in addition to a segmented image. Then, this output is combined with the original image in the hazard analysis and avoidance model. In this study, we develop a hazard analysis and avoidance model based on the following equation:

$$I = (1 - T_f) \times I_{original} + T_f \times I_{segmented}$$

where $I$ is the image used to predict the autonomous vehicle driving model; $I_{original}$ is the data from the center dashboard camera of the vehicle; $I_{segmented}$ is the segmented image of $I_{original}$ containing the hazardous object detected and segmented; and $T_f$ is the threat value of the detected hazardous object or physical-world threat object. This threat value depends on the position of a detected object on a driving lane. If the object is dangerous to the autonomous vehicle, it will have a high threat factor, while a negligible threat object will have a lower threat factor. This threat value depends on the longitudinal and latitudinal distance from the autonomous vehicle. Depending on the hazardous object localization technique, we have used two procedures to determine the threat value: (ii) Procedure 1 - threat value determination using a distance measurement sensor (e.g., Radar); and (ii) Procedure 2 – threat value determination using image segmentation.

*Procedure 1 - threat value determination using a distance measurement sensor*
According to the first procedure, we measure the longitudinal distance ($l_x$), and latitudinal distance ($l_y$) of hazardous objects from the vehicle using a distance measurement sensor. If the vehicle is moving forward (longitudinal movement) or steering towards (latitudinal movement) the hazard the value of $l_x$ and $l_y$ decreases, respectively, and hence the hazard poses a higher threat of colliding with the vehicle. We consider the hazard as a threat to the vehicle if the hazard is within the longitudinal distance, $l_{x,max}$ and latitudinal distance, $l_{y,max}$. In our study, we use the Radar sensor to measure the longitudinal distance and the latitudinal distance, and we measure the threat value using the following equations:

$$T = \sqrt{\left(\frac{l_{x,max} - l_x}{l_{x,max}}\right)^2 + \left(\frac{l_{y,max} - l_y}{l_{y,max}}\right)^2}$$

$$T_f = \begin{cases} \dfrac{T - T_{min}}{T_{max} - T_{min}} & if\ l_y \leq\ l_{y,max}\ ; and\ l_x \leq\ l_{x,max} \\ 0 & if\ l_y >\ l_{y,max}\ ; or\ l_x >\ l_{x,max} \end{cases}$$

where, $T_f$ is the threat value corresponding to the hazardous object; $l_x$ and $l_y$ are the longitudinal distance and latitudinal distance in centimeters (cm) to the detected hazard from the vehicle, respectively; $l_{x,max}$ and $l_{y,max}$ is the maximum longitudinal distance and maximum latitudinal distance, correspondingly, to consider the hazard as a threat; and $T$ is the threat value calculated from the longitudinal and the latitudinal distance. Then, the value of $T$ is normalized using the Min-Max normalization technique to obtain a value between 0 to +1 to determine the final threat value, $T_f$ *(35)(36)*. In our experiment, we have selected $l_{x,max}$ as 6000cm as this is the Radar's maximum range of finding an object in our experimental setup, and $l_{y,max}$ is selected as 370cm which is the standard lane width of a roadway. We can visualize the relationship between the threat value, and longitudinal and latitudinal distance in Figure 3.

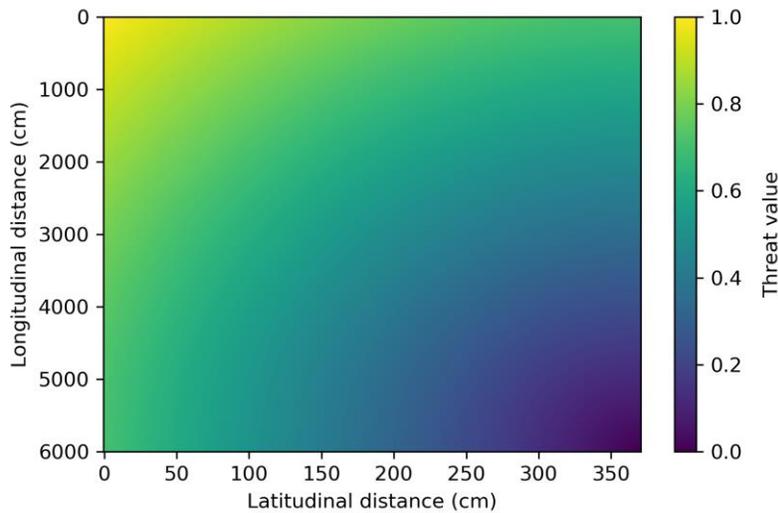

**FIGURE 3 Heatmap for the threat value based on the longitudinal and latitudinal distance of a hazardous object using Radar sensor data.**

*Procedure 2 – threat value determination using image segmentation*
In this procedure, instead of using a Radar sensor, we use the segmented image to calculate the threat value. In this way, we can eliminate the use of any sensor data besides camera video feed.

After the image segmentation, we get the image coordinates $(x, y)$ of the hazard. As the camera is located at the center dashboard of the vehicle facing the front roadway, we measure the relative distance of the hazardous object in the image of size $(h, w)$, from the bottom center pixel, $(h, \frac{w}{2})$ to quantify the threat. We calculate the threat based on the location of the hazard in the segmented image using the following equation:

$$T_f = 1 - \sqrt{\frac{(x - h)^2 + \left(y - \frac{w}{2}\right)^2}{h^2 + \left(\frac{w}{2}\right)^2}}$$

where, $T_f$ is the threat value corresponding to the hazardous object located in the segmented image at location $(x, y)$ pixels, where $(x, y)$ is the pixel value closest to the bottom center pixel, $(h, \frac{w}{2})$, of the image. The value of $h$ and $w$ indicates the height and width of the image, respectively. As the camera of the vehicle is located at the center of the vehicle facing the front roadway, we subtract $h$ and $\frac{w}{2}$ values from the $x$ and $y$ values, respectively, to obtain the longitudinal and latitudinal distance of the hazard relative to the front center of the vehicle. As we described the equation above, we calculate the threat value. We can visualize the threat value in Figure 4, where the threat value decreases as the object moves from the center bottom pixel of the image.

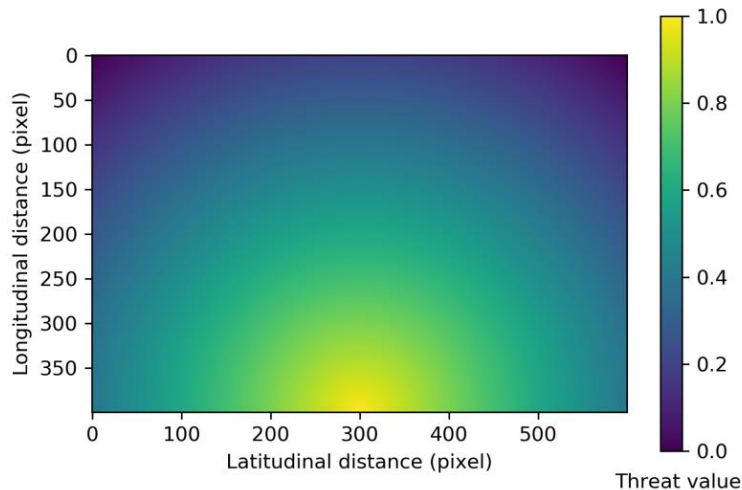

**FIGURE 4 Heatmap for the threat value based on the location of hazard using pixel value from the segmented image.**

### DNN-based autonomous vehicle driving model
In our study, we have implemented an autonomous vehicle driving model similar to DAVE-2, an end-to-end autonomous vehicle driving model (*7*). As shown in Figure 5, the network receives an input image of 400x600x3 pixels and produces a steering wheel angle as an output. This network includes one lambda layer, one normalization layer, five convolution layers (Conv2D), and four fully connected (FC) layers. We have used a 5x5 kernel (i.e., filters) and 2x2 stride (i.e., the increment of kernel movement) in the first 3 Conv2D layers, and a 1x1 stride and a 3x3 kernel in the last two Conv2D layers. The entire network contains a total of 7,970,619 trainable parameters.

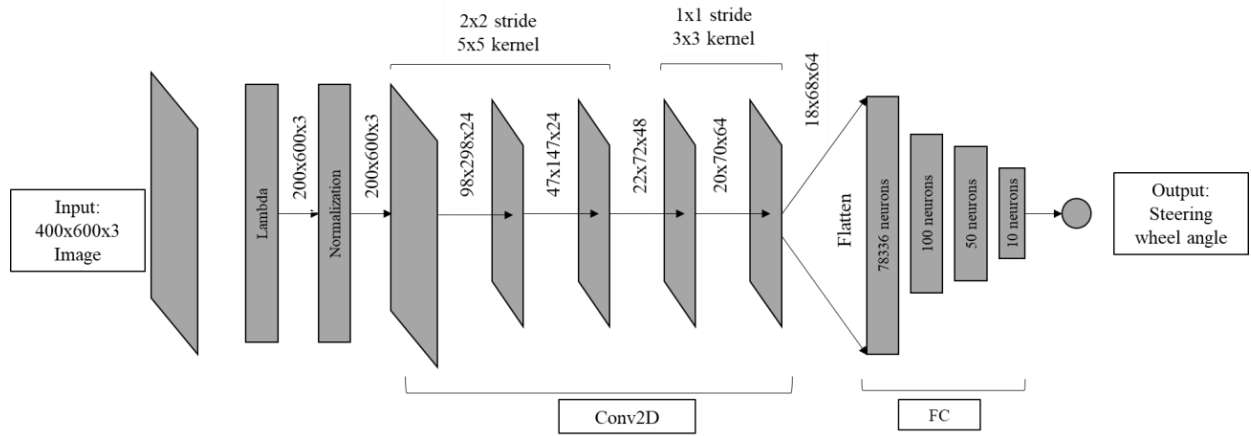

**FIGURE 5 CNN-based end-to-end autonomous vehicle driving model used in this study.**

We train our driving model of an autonomous vehicle from the output of hazard analysis and avoidance model followed by the deployment to test the performance. After the training, our trained autonomous vehicle driving model is aware of hazardous objects on the roadway and produces a steering wheel angle to navigate safely around the hazard.

**EXPERIMENTAL SETUP**

In the experimental setup, we describe the data collection method, data preparation, and data augmentation; and finally, we train and validate the DNN-based autonomous vehicle driving model. The steps of our experiment setup are as follows:

**Data Collection**

For this study, we have used the robotics simulation platform Webots (*37*) to create the roadway environment with hazardous objects and to collect the data including the driving attributes of the camera image, timestamp, location, vehicle speed, and steering wheel angle. The following subsections describe the collection procedure of the dataset.

*Roadway Environment Setup*

The roadway built in the simulation consists of two lanes in each direction and 1663m in length with 16 curves (having 45 degree to 90 degree radius of curvature) and two intersections as shown in Figure 6(a). Six additional non-autonomous vehicles are placed randomly on the roadway. The hazardous debris, which includes five objects: rocks, wooden boxes, oil barrels, wooden pallets, and sections of pipe are created in Webots (*37*) and placed randomly on the roadway as shown in Figure 6(b).

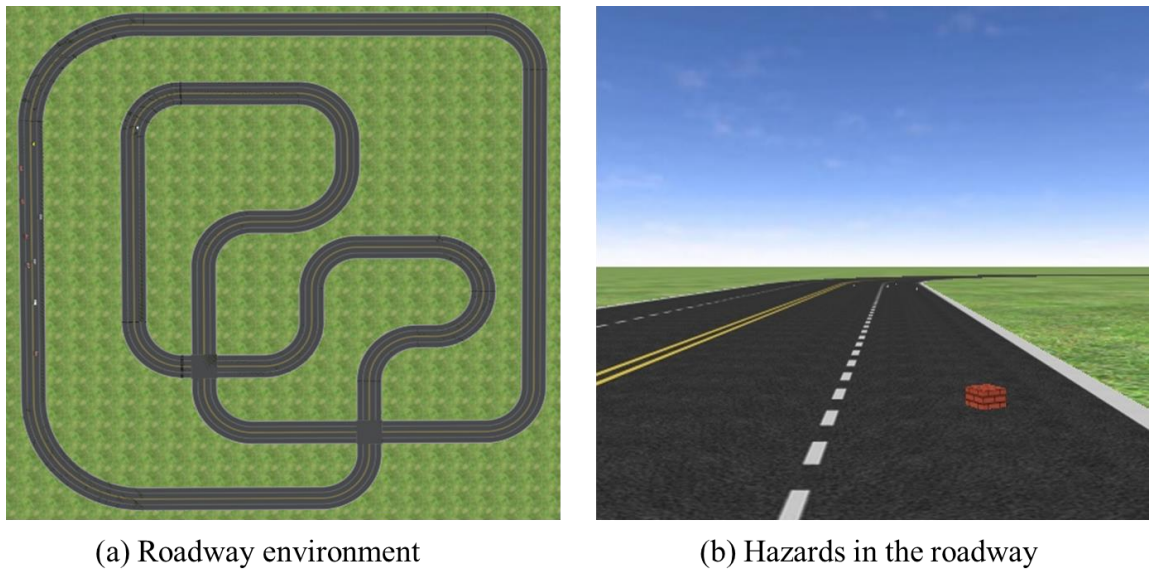

(a) Roadway environment                          (b) Hazards in the roadway

**FIGURE 6 Roadway environment setup for an autonomous vehicle with hazardous objects.**

*Autonomous Vehicle Setup*

For collecting the data, the autonomous vehicle is equipped with three dashboard cameras, a front, left and right camera (as shown in Figure 7) and a Radar sensor. The data collected using these cameras are used to train the end-to-end autonomous vehicle driving model. For example, as seen in Figure 8, the images collected by the left and right camera differ from the center camera. After training, the autonomous vehicle uses only a single front camera to navigate through the roadway, similar to the DAVE-2 system (*7*). In our developed driving model, we have used the Delphi ESR Radar sensor, which is commercially used in the existing autonomous vehicles (*38*). We have used the medium range mode configurations (horizontal field of view of 90 degrees and a maximum range of 6000 cm) of the Radar sensor in our autonomous vehicle (*39*). We have also equipped the vehicle with three other Radar sensors in three directions (left, right and back side) for monitoring the near-by traffic condition and vehicles. These Radars sensors are also configured in the medium range mode.

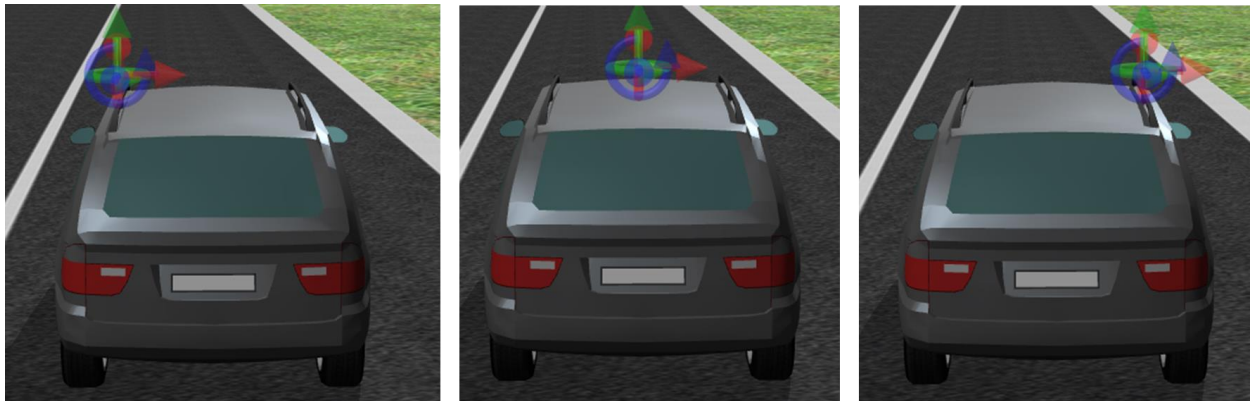

**FIGURE 7 Camera placements in the autonomous vehicle (left, center, and right cameras).**

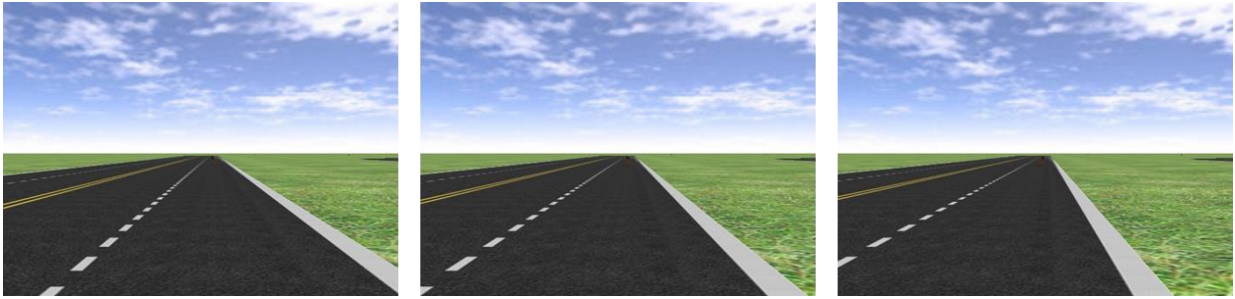

**FIGURE 8 Example of images collected by the three cameras in the autonomous vehicle (left, center, and right camera images (from left to right))**

## Data Preparation

After collecting the data, we prepare the image dataset for training the end-to-end driving model by normalizing and resizing. As shown in Figure 9, the steering wheel angle output is normalized between the values of -0.5 and +0.5, where a positive value indicates the steering to the right, and a negative value represents steering to the left using linear transformation following this equation:

$$\theta_{normalized} = -0.5 + \max\left(0, \min\left(1.0, \frac{\theta_{raw} - \theta_{min}}{\theta_{max} - \theta_{min}}\right)\right)$$

where, $\theta_{normalized}$ is the normalized steering angle between -0.5 and +0.5; $\theta_{raw}$ is the actual steering wheel angle (in radians) measured from the vehicle; $\theta_{max}$ and $\theta_{min}$ are the maximum and minimum steering wheel angle, respectively. We also normalize the input images for training, which is necessary to improve the DNN model performance (*40*). Normalization is also done on the input images. The red, green, and blue (RGB) channel values of the input images are normalized between the values of -1.0 and +1.0, and their top 200 pixels is cropped using a Lambda layer (as shown Figure 10) as top portion of the image is not necessary to predict the steering wheel angle, and doing so does not impact the steering wheel angle output of the driving model. For all data collected, we use an online image annotation tool, LabelMe (*41*), for labeling the hazardous object and segmented image. Using this tool, we have created the ground truth data for training the image segmentation model for detecting and segmenting the hazards in an image.

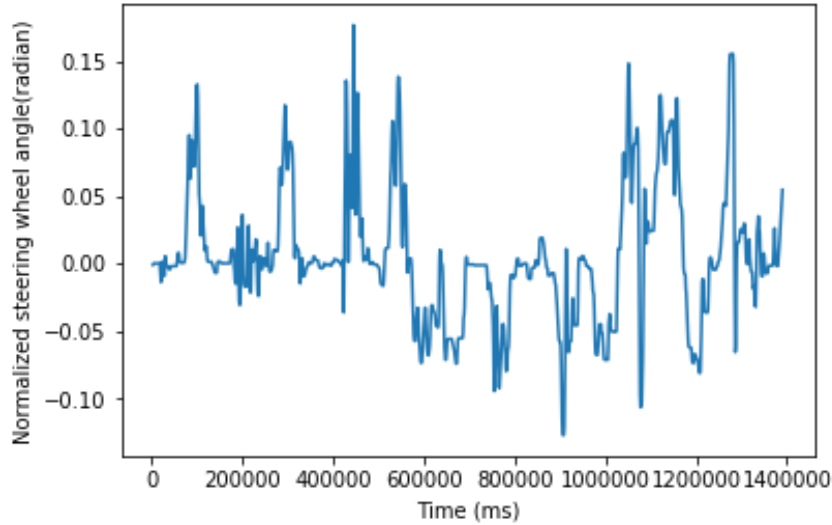

**FIGURE 9 Example of a normalized steering wheel angle plot from the training dataset.**

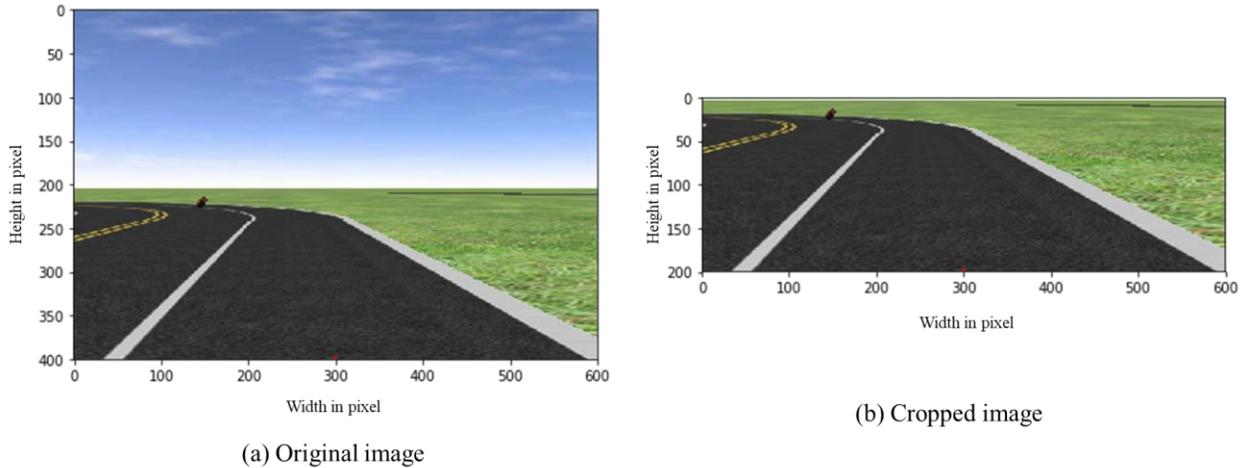

(a) Original image                                    (b) Cropped image

**FIGURE 10 Example of an original and cropped image in the training dataset.**

## Data Augmentation

To obtain satisfactory performance from the driving model, it is necessary to train the model on multiple training datasets. Using the techniques of data augmentation, we have created additional data from the existing data through affine transformation (*42*), specifically random rotation, random brightness change, and horizontal flipping of the images, to double the size of the dataset as shown in Table 1. From our first simulation, we have collected 1390 images in total, and we have split the image dataset in training (i.e., 1112 images) and validation dataset (i.e., 278 images) as shown in column 2 of Table 1. Then we have doubled the dataset size (i.e., 2780 images) using data augmentation as presented in column 3 of Table 1. Among these 2780 images, 2224 images are used for training, and the remaining 556 images are used for validation. Among the 2224 images used for training, 468 images contained hazards. Furthermore, we have collected 104 images from a second simulation where all the images contained hazard. These 104 images are used to evaluate or test the driving model performance.

**TABLE 1 Dataset description**

| Dataset type | Collected dataset size | Dataset size after data augmentation |
|---|---|---|
| **Dataset size** | 1390 | 2780 |
| **Training dataset size** | 1112 | 2224 |
| **Validation dataset size** | 278 | 556 |
| **Testing dataset size (all containing hazard)** | 52 | 104 |

**Model Training and Validation**

After the development of the end-to-end autonomous vehicle driving model, we train it using the augmented dataset. This dataset is divided into two, 80% in a training set (2224 images as per Table 1) and the remaining 20% in a validation set (556 images as per Table 1). We then train three models for our evaluation:

- *Case 1:* A model trained on a dataset that includes hazards but without considering them as a separate input feature**.**

- *Case 2:* A model trained on a dataset that considers hazards as separate input features and uses a distance measurement sensor and image segmentation. In this case, the threat value is determined using a distance measurement sensors (Radar in our case), following the Procedure 1 as described in the method section.

- *Case 3:* A model trained on a dataset that considers hazards as separate input features and uses image segmentation. In this case, the threat value is determined using the image segmentation, following the Procedure 2 as described in the method section.

For the training of the autonomous vehicle driving model, we have used Adam optimizer that can change the learning rate dynamically(*43*). The mean square error based loss function, a dropout rate of 0.5 in the last four FC layers, and L2 regularization are used to reduce overfitting and under-fitting and to minimize training error (*44*). We use model checkpoints to stop the training when the validation loss is not decreasing over time (*45*). Figure 11(a) shows the performance of the model training for Case 1, where the training is stopped after 14 epochs because the model does not exhibit much improvement after 11 epochs. We observe no overfitting or under-fitting during the training. In Case 2, the model stopped training after 16 epochs (as shown in Figure 11(b)), and in Case 3, the model stopped the training after 15 epochs (as shown in Figure 11(c))

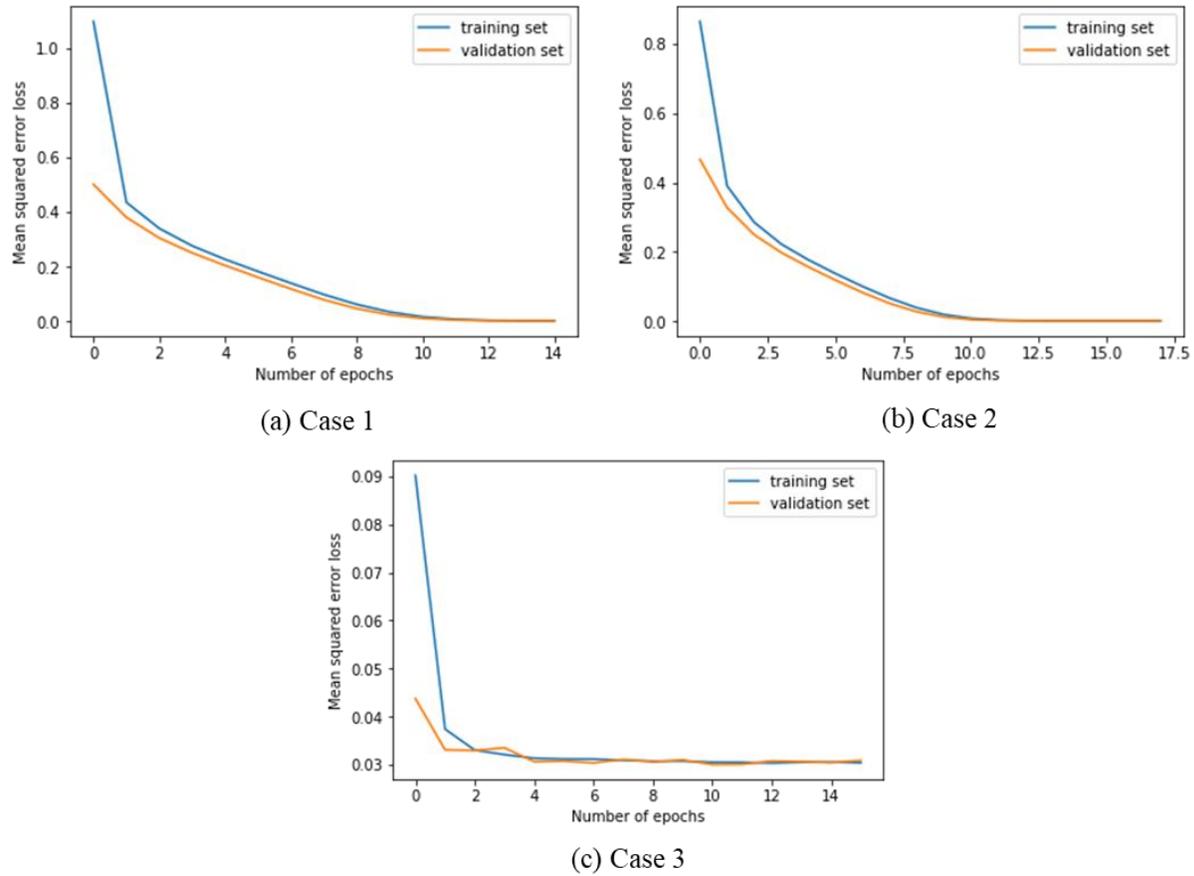

(a) Case 1                                                          (b) Case 2

(c) Case 3

**FIGURE 11 Training and validation performance of the end-to-end driving model for Case 1, Case 2, and Case 3 on the training and validation dataset.**

## ANALYSIS RESULTS

After training and validating the model using the dataset from the first simulation, we evaluate the trained end-to-end autonomous vehicle driving model using the test dataset of 104 images (as depicted in Table 1). We have created this dataset of 104 images from a second simulation where all debris are placed in the middle of the driving lane, and we measure the predicted steering wheel angle for each test image. In this second simulation, first, we create the ground truth by manually driving the vehicle on the roadway. Then we deploy the trained end-to-end autonomous vehicle driving model for Case 1, Case 2 and Case 3. We then analyze the performance of the model for each cases using the following quantitative measures: root mean square error (RMSE) and mean absolute error (MAE), and a qualitative measure through visualization.

### Quantitative Results of Model Performance

The quantitative results include the RMSE and the MAE, are measured by comparing the predicted steering wheel angle with the actual steering wheel angle (i.e., ground truth data). We define the RMSE and MAE as follows:

$$RMSE = \sqrt{\frac{1}{N}\sum_{i=1}^{N}(G_i - P_i)^2}$$

$$MAE = \frac{1}{N}\sum_{i=1}^{N}(|G_i - P_i|)$$

where N is the total number of images in the testing dataset; and $G_i$ and $P_i$ are the ground truth and predicted steering wheel angle, respectively, for the $i^{th}$ image of the testing dataset. As shown in Figure 12, both the RMSE and MAE are higher for Case 1 than Case 2 and Case 3. A lower RMSE and MAE indicate that the predicted steering wheel angle is closely following the actual steering wheel angle or ground truth data related to steering wheel angle.

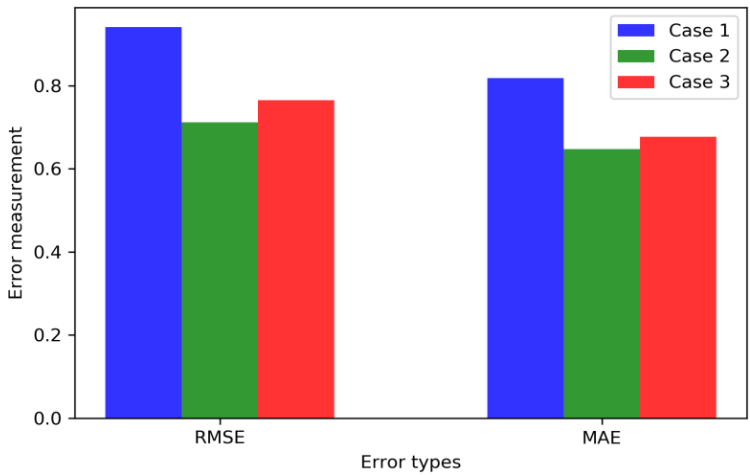

**FIGURE 12 Error measurement on the testing dataset.**

We measure the steering wheel angle prediction accuracy and improvement of Case 2 and Case 3, over Case 1. By comparing $RMSE$ of Case 2 and Case 3 with $RMSE_{case1}$, we calculate the steering wheel angle prediction improvement based on the equation below:

$$Percentage\ of\ improvement = \left(\frac{|\ RMSE - RMSE_{case1}|}{RMSE_{case1}} \times 100\right)\ \%$$

Based on our experiment, we found a 21% improvement in the steering wheel angle prediction of Case 2 over Case 1, and 18% improvement in the steering wheel angle prediction of Case 3 over Case 1. The results suggest that both Case 2 and Case 3 improve the autonomous vehicle navigation to avoid an unexpected hazard on the roadway.

**Qualitative Results for Driving Direction**
Figure 13 shows the qualitative results of our study on the autonomous vehicle driving direction. To obtain the qualitative measurement, we transform the steering wheel angle (-0.5 to +0.5) into a driving direction angle (-25 degrees to +25 degrees) using linear transformation. In Webots, the steering wheel angle follows the Ackermann geometry, representing a linear relationship between steering wheel angle and driving direction (*37*)(*46*). The prediction accuracy can be presented qualitatively by observing the driving direction angle or angle of movement of the autonomous vehicle. For example, Figure 13 shows that the continuous steering wheel output of data from the time step of 64000 milliseconds (ms) to 72000ms window for ground truth, Case 1, Case 2, and Case 3. In the presence of a hazard on the roadway, the autonomous vehicle driving model is producing the output for maneuvering the autonomous vehicle. According to Figure 13, the autonomous vehicle is moving towards the left for each case. For example, in Case 1, at time step 66000ms, the predicted driving direction is +5.2 degrees, causing the car to move closer to the hazard (represented here as a box) compared to Case 2 and Case 3. However, in Case 2 and Case 3, the predicted driving direction is +11.7 degree and +9.28 degree, respectively, which is a value closer to the ground truth than in Case 1. Overall, the qualitative results indicate better accuracy prediction for Case 2 and Case 3 than for Case 1.

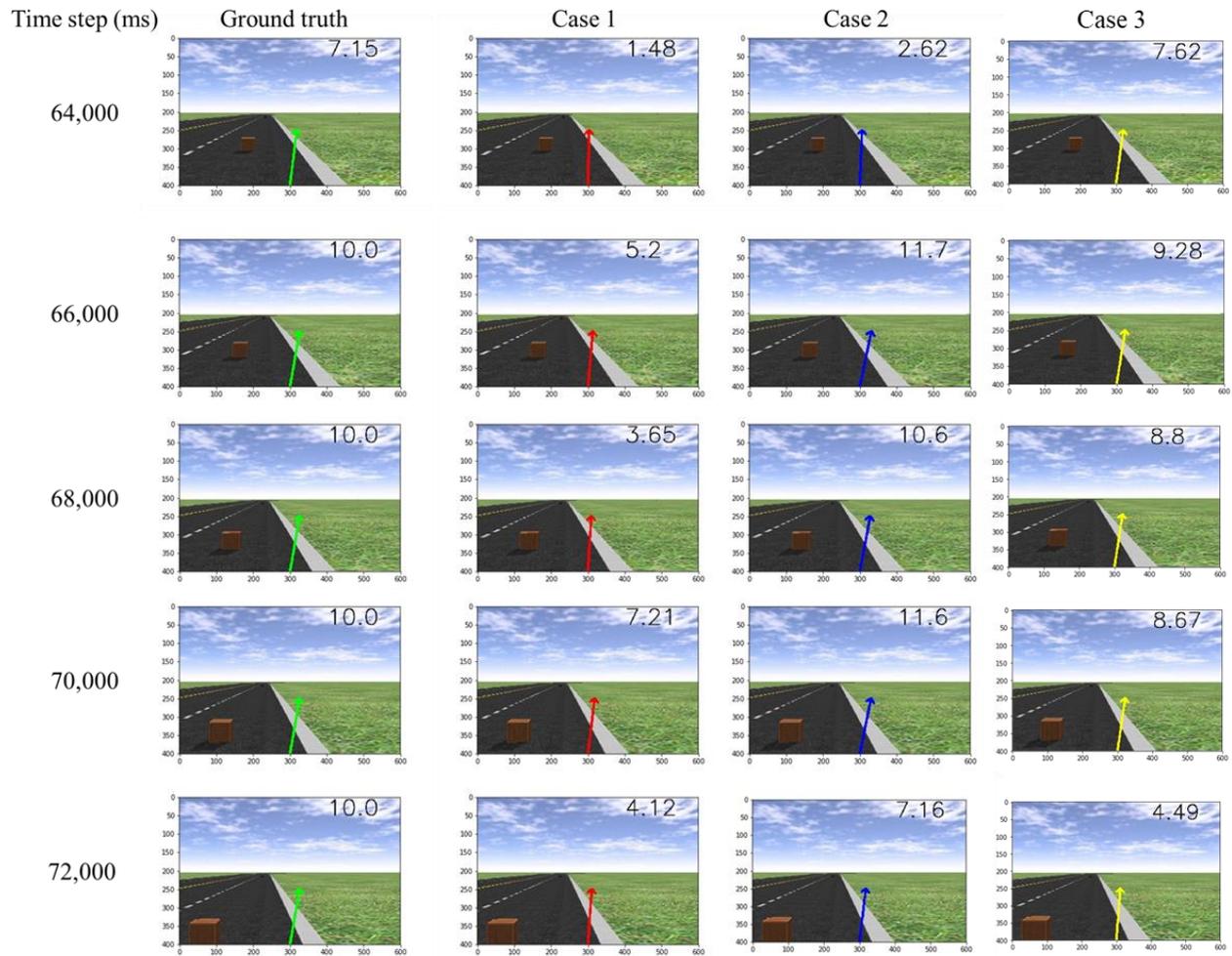

**FIGURE 13 Qualitative results of ground truth, Case 1, Case 2, and Case 3 of the driving direction.**

**Quantitative Results for Driving Direction**

Following the Frenet coordinate system, we have performed quantitative analyses of hazard avoidance. In a Frenet coordinate system, the longitudinal movement and latitudinal movement are represented in x-axis and y-axis, respectively (*47*). Instead of following the Frenet coordinate system for the performance evaluation, we have plotted the time step in the x-axis and latitudinal movement in the y-axis (see Figure 14) to show the deviation of latitudinal movement of an autonomous vehicle and how the vehicle avoids a hazardous object for different cases (as described in the 'Model Training and Validation' subsection) over the time. We analyze the trajectory of the autonomous vehicle and calculate the RMSE between the vehicle trajectory of each case and the ground truth. In Figure 14, we present the autonomous vehicle trajectories for all three cases from the time step 62000ms to 72000ms to show how accurately the vehicle following the ground truth trajectory data for each case to avoid the hazardous object. For Case 2, the vehicle trajectory produced from the autonomous vehicle driving systems is closely following the ground truth vehicle trajectory compared to Case 1 and Case 3. However, in all cases, i.e., Case 1, Case 2, and Case 3, the vehicle is able to avoid the hazard (Figure 14). In Case 1, the RMSE value was 0.52. On the other hand, the RMSE values for Case 2 and Case 3 are 0.07 and 0.23, respectively. We perform a statistical significance test (pairwise t-test) between the ground truth and each case separately at a 95% confidence interval. We find that Case 1 is significantly different from the ground truth at a 95% confidence interval. However, Case 2, which uses both image segmentation and a distance measurement sensor, and Case 3, which only uses the segmented image, are not significantly different from the ground truth. Thus, based on the statistical analyses of Case 2 and Case 3, we achieve the same level of performance using image segmentation, and not using any additional distance measurement sensor, i.e., Radar.

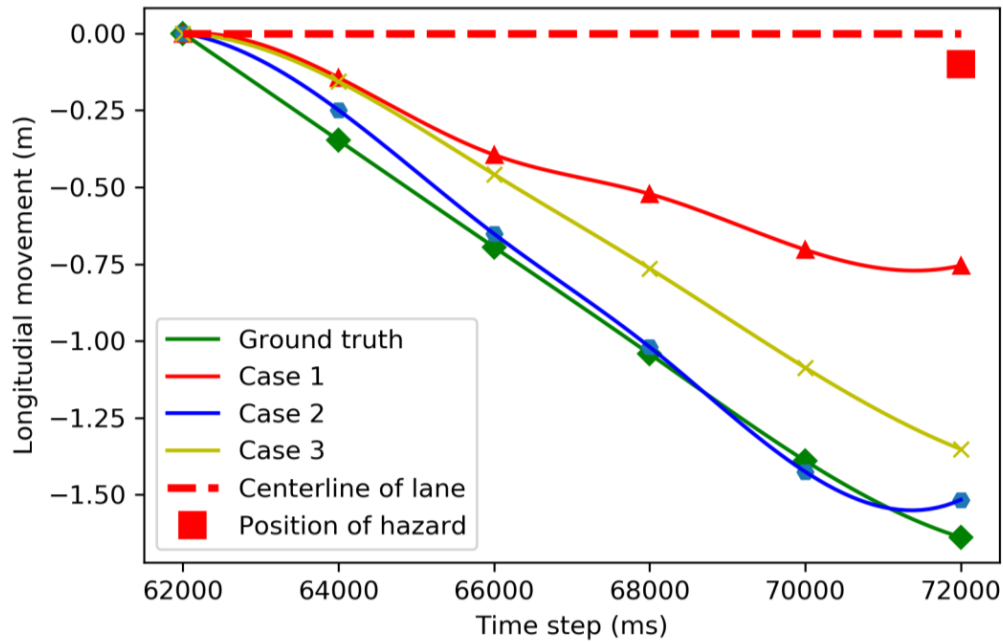

**FIGURE 14 Trajectory of the autonomous vehicle for ground truth, Case 1, Case 2, and Case 3.**

## CONCLUSIONS

Detecting unexpected hazards on a roadway is a crucial task for the safe operation of an autonomous vehicle. In this work, we have developed and evaluated a DNN-based driving system for autonomous vehicles in an unexpected hazardous roadway environment. First, we detect the hazard, and then using semantic segmentation, we extract the hazard information and perform data fusion to improve the navigation of an autonomous vehicle. This study makes the following contributions to the current body of research: (i) we evaluate the effect of the hazardous roadway environment on the DNN-based driving system of an autonomous vehicle; (ii) we develop a DNN-based driving system for autonomous driving that can address an unexpected hazardous roadway environment and can navigate the autonomous vehicle safely through this environment. More specifically, we explore the object detection and semantic segmentation based deep learning models to address an unsafe navigational problem; (iii) we contribute a new dataset that can be used by the autonomous vehicle community to improve the driving model in unexpected hazardous roadway environment. Based on the analyses result, we conclude that our method improved the safety of the autonomous vehicle by 21% in terms of avoiding hazards, compared to a vision-based navigation system of autonomous vehicles having no hazard detection and segmentation as separate input features. Future work will include fusing the temporal and spatial information into the DNN-based model, potentially further improving the safety of autonomous vehicles operating in an unexpected hazardous roadway environment.

## ACKNOWLEDGMENTS

This paper is based on the study supported by the USDOT Center for Connected Multimodal Mobility (Tier 1 University Transportation Center) Grant headquartered at Clemson University. Any opinions, findings, and conclusions or recommendations expressed in this paper are those of the author(s) and do not necessarily reflect the views of the USDOT Center for Connected Multimodal Mobility ($C^2M^2$). The U.S. Government assumes no liability for the contents or use thereof. The authors would like to thank Aniqa Chowdhury and Mizanur Rahman, Ph.D. for editing the paper.

## AUTHOR CONTRIBUTION STATEMENT

The authors confirm contribution to the paper as follows: study conception and design: M. Islam, M. Chowdhury, H. Li, and H. Hu; Data collection: M. Islam; Analysis and interpretation of results: M. Islam, M. Chowdhury, H. Li; draft manuscript preparation: M. Islam, M. Chowdhury, H. Li, and H. Hu. All authors reviewed the results and approved the final version of the manuscript.